\documentclass[conference]{IEEEtran}
\IEEEoverridecommandlockouts
\usepackage{cite}
\usepackage{amsmath,amssymb,amsfonts}
\usepackage{algorithmic}
\usepackage{algorithm}
\usepackage{array}
\usepackage{graphicx}
\usepackage{textcomp}
\usepackage{xcolor}
\usepackage{adjustbox}
\usepackage[caption=false,font=normalsize,labelfont=sf,textfont=sf]{subfig}
\usepackage{stfloats}
\usepackage{url}
\usepackage{verbatim}
\usepackage{float}
\usepackage{booktabs}
\usepackage{multirow}

\def\BibTeX{{\rm B\kern-.05em{\sc i\kern-.025em b}\kern-.08em
    T\kern-.1667em\lower.7ex\hbox{E}\kern-.125emX}}

\begin{document}

\title{Temporal-Aware Graph Attention Network for Cryptocurrency Transaction Fraud Detection}

\author{\IEEEauthorblockN{1\textsuperscript{st} Zhi~ZHENG}
\IEEEauthorblockA{\textit{School~of~Finance~and~Business} \\
\textit{Shanghai~Normal~University}\\
Shanghai, China \\
Email:hhmcn@outlook.com}
\and
\IEEEauthorblockN{2\textsuperscript{nd} Bochuan ZHOU}
\IEEEauthorblockA{\textit{School~of~Finance~and~Business} \\
\textit{Shanghai~Normal~University}\\
Shanghai, China \\
Email:1000553566@smail.shnu.edu.cn}
\and
\IEEEauthorblockN{3\textsuperscript{rd} Yuping SONG}
\IEEEauthorblockA{\textit{School~of~Finance~and~Business} \\
\textit{Shanghai~Normal~University}\\
Shanghai, China \\
Email:songyuping@shnu.edu.cn}
}

\maketitle

\begin{abstract}
Cryptocurrency transaction fraud detection faces the dual challenges of increasingly complex transaction patterns and severe class imbalance. Traditional methods rely on manual feature engineering and struggle to capture temporal and structural dependencies in transaction networks. This paper proposes an Augmented Temporal-aware Graph Attention Network (ATGAT) that enhances detection performance through three modules: (1) designing an advanced temporal embedding module that fuses multi-scale time difference features with periodic position encoding; (2) constructing a temporal-aware triple attention mechanism that jointly optimizes structural, temporal, and global context attention; (3) employing weighted BCE loss to address class imbalance. Experiments on the Elliptic++ cryptocurrency dataset demonstrate that ATGAT achieves an AUC of 0.9130, representing a 9.2\% improvement over the best traditional method XGBoost, 12.0\% over GCN, and 10.0\% over standard GAT. This method not only validates the enhancement effect of temporal awareness and triple attention mechanisms on graph neural networks, but also provides financial institutions with more reliable fraud detection tools, with its design principles generalizable to other temporal graph anomaly detection tasks.
\end{abstract}


\section{Introduction}
In the digital finance era, transaction fraud has emerged as a critical threat to global financial system security. According to the Nilson Report \cite{nilson2024}, global credit card fraud losses reached \$33.8 billion in 2023. The cryptocurrency domain presents even more severe challenges—Chainalysis \cite{chainalysis2024} reports that \$40.9 billion in cryptocurrency flowed to illicit addresses in 2024, with money laundering, Ponzi schemes, and ransomware accounting for over 70\%. These fraudulent activities not only cause substantial economic losses but also severely undermine public trust in digital financial systems.

Early fraud detection research predominantly relied on rule-based systems and traditional machine learning approaches. Leonard et al.~\cite{LEONARD1995350} proposed a rule-based expert system for credit card fraud early warning. Subsequently, researchers introduced machine learning to this field, including Support Vector Machines (SVM)~\cite{JSJZ201108091}, Multilayer Perceptrons (MLP)~\cite{Elmi2013DetectingSB}, and Generative Adversarial Networks (GAN)~\cite{1020727280.nh}. However, these methods exhibit several limitations when processing financial transaction data: First, feature extraction relies heavily on domain expert knowledge, hindering automatic discovery of latent patterns within data; second, treating transactions as independent samples neglects inter-transaction relationships; third, performance degrades significantly under class imbalance conditions, while severe positive-negative sample ratio disparity is ubiquitous in fraud detection tasks \cite{he2009learning}.

Recently, Graph Neural Networks (GNNs) have provided novel technical pathways for transaction fraud detection by modeling transactions as graph structures, enabling simultaneous utilization of node features and network topology information \cite{KipfW16,velivckovic2018graph}. Qin et al. \cite{JSJA2024S2123} first applied GNNs to banking transaction fraud detection; Duan et al. \cite{duan2024catgnnenhancingcreditcard} proposed CaT-GNN incorporating causal invariant learning; Li et al. \cite{singh2024heterogeneousgraphautoencodercreditcard} designed heterogeneous graph autoencoders to enhance minority class identification. Despite these advances, several research gaps persist: (1) existing methods inadequately utilize transaction temporal information, while they play crucial roles in fraud pattern identification \cite{rossi2020temporal,xu2020inductive}; (2) attention mechanism designs remain relatively simplistic, struggling to comprehensively consider local and global information; (3) systematic solutions for extreme class imbalance are lacking.

Based on this analysis, this paper proposes Augmented Temporal-Aware Graph Attention Network (ATGAT) for cryptocurrency transaction fraud detection. The main contributions include: (1) designing a temporal embedding module incorporating multi-scale time difference features and periodic encoding; (2) proposing a temporal-aware triple attention mechanism learning node representations from structural, temporal, and global dimensions; (3) employing a weighted cross-entropy loss function to mitigate class imbalance issues. Experimental results on the Elliptic++ dataset \cite{elmougy2023demystifying} demonstrate the proposed method's superiority over existing approaches across primary evaluation metrics.

\section{Method}
\subsection{Problem Formalization}

Let $G = (V, E, X, Y, T)$ denote a directed cryptocurrency transaction graph, where:
\begin{itemize}
    \item $V = \{v_i\}_{i=1}^{N}$ represents the transaction node set, with each node corresponding to a Bitcoin transaction;
    \item $E \subseteq V \times V$ denotes the directed edge set, where an edge $(i \to j)$ exists if transaction $v_i$ sends Bitcoin to $v_j$;
    \item $X \in \mathbb{R}^{N\times d}$ is the node feature matrix, where each node has $d$-dimensional raw features (including transaction amount, time step information, statistical features, etc.);
    \item $Y \in \{0,1\}^{N}$ represents node labels, where 0 indicates legitimate transactions and 1 indicates fraudulent transactions. The training set contains labels for only a subset of nodes, with remaining nodes requiring prediction;
    \item $T = \{t_i\}_{i=1}^{N}$ is the node timestamp sequence, where $t_i$ denotes the time when the $i$-th transaction was broadcast to the network (indexed by discrete time steps, e.g., 1 to 49).
\end{itemize}

We define the supervised node classification task: given a transaction graph $(V, E, X, T)$ with partially known labels, including labeled node set $V_\text{labeled} \subset V$ partitioned into training set $V_\text{train}$ and validation set $V_\text{val}$, learn a mapping function
\begin{equation}
f_\theta: (V, E, X, T) \rightarrow \hat{Y},
\end{equation}

such that accurate predictions $\hat{y}_i\in\{0,1\}$ can be generated for validation set nodes.

\subsection{Model Architecture}\label{sec:enhanced-gat}

\begin{figure}[htbp]
    \centering
    \includegraphics[width=0.48\textwidth]{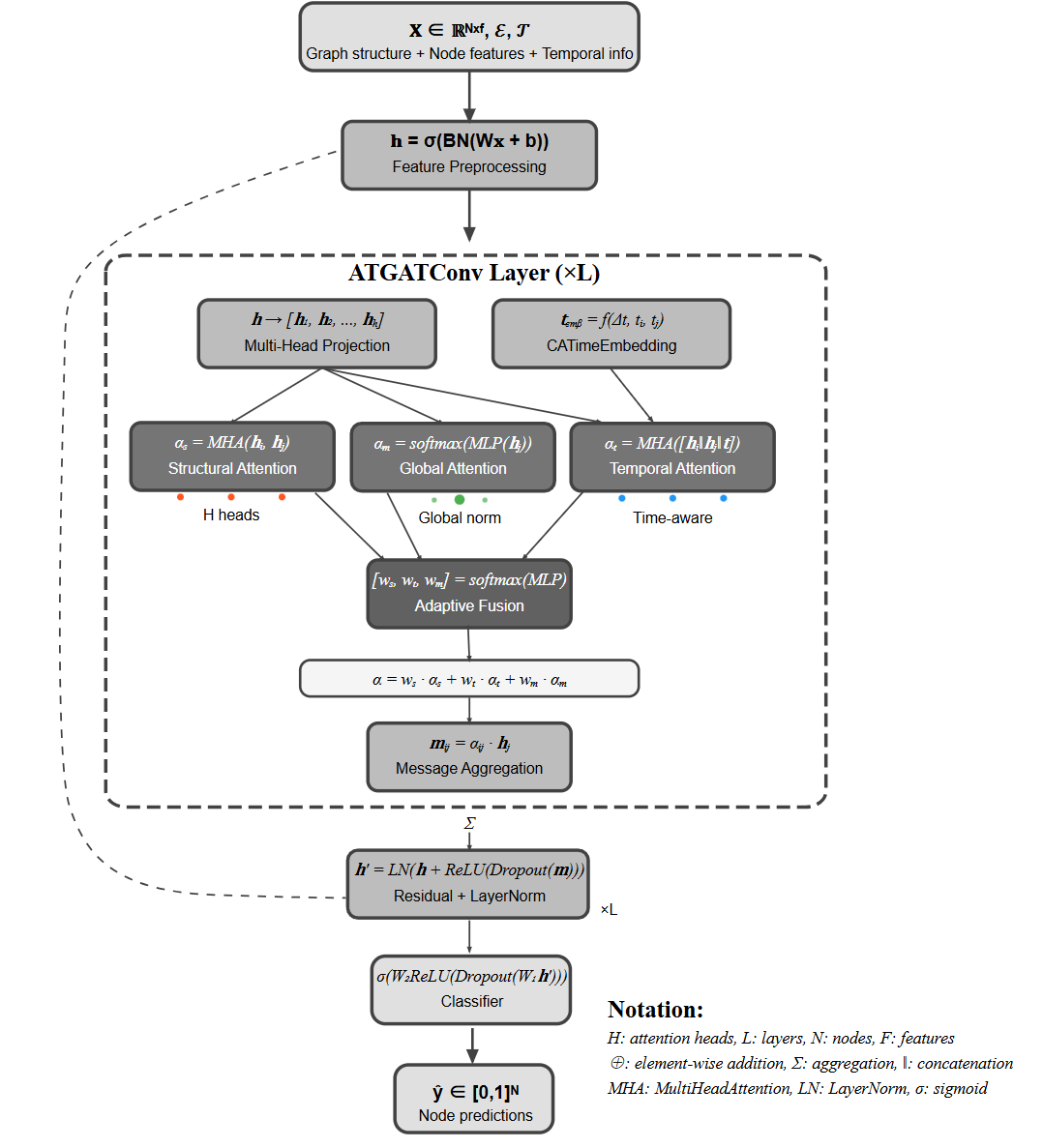}
    \caption{Overall architecture of ATGAT.}
    \label{fig:architecture}
\end{figure}

Addressing the temporal characteristics and class imbalance issues of cryptocurrency transaction graphs, this paper proposes the ATGAT architecture (Fig.~\ref{fig:architecture}). The architecture achieves synergy through three core innovative modules: advanced temporal embedding captures multi-scale temporal patterns, temporal-aware triple attention mechanism fuses information from different dimensions, and weighted loss function handles data imbalance.

\subsection{Core Modules}\label{sec:module-detail}
\subsubsection{Advanced Temporal Embedding}
\label{subsec:time-embedding}

To comprehensively capture multi-level temporal relationships in cryptocurrency transactions, this paper designs a temporal embedding module comprising three components:

\paragraph{Basic Time Difference Projection}
For node pair $(i,j)$ with time difference $\Delta t_{ij} = |t_i - t_j|$, we map through a linear layer to $d_1$ dimensions:
\begin{equation}
  \mathbf{f}_{\Delta}^{(ij)} = \mathbf{W}_{\Delta}\,\bigl[\Delta t_{ij}\bigr] + \mathbf{b}_{\Delta}
  \in \mathbb{R}^{d_1},
\end{equation}
where $\mathbf{W}_{\Delta}\in\mathbb{R}^{d_1\times1}$, $d_1 = \lfloor d_t/4\rfloor$, and $d_t$ is the final temporal embedding dimension.

\paragraph{Multi-scale Temporal Feature Extraction}
We extract three scales of nonlinear transformation features: $f_{\mathrm{short}}(\Delta t_{ij}) = \Delta t_{ij}$, $f_{\mathrm{medium}}(\Delta t_{ij}) = \log(\Delta t_{ij} + 1)$, $f_{\mathrm{long}}(\Delta t_{ij}) = \sqrt{\Delta t_{ij} + 1}$. After concatenation, these are mapped to $d_3 = \lfloor d_t/4\rfloor$ dimensions through a linear layer:
\begin{equation}
  \mathbf{f}_{\mathrm{ms}}^{(ij)} = \mathbf{W}_{\mathrm{ms}}\,[f_{\mathrm{short}}, f_{\mathrm{medium}}, f_{\mathrm{long}}]^T + \mathbf{b}_{\mathrm{ms}} \in \mathbb{R}^{d_3}.
\end{equation}

\paragraph{Periodic Position Encoding}
Inspired by Transformer\cite{vaswani2023attentionneed}, we construct $d_{\mathrm{pos}}$-dimensional sinusoidal position encoding for each node's timestamp $t_i$:
\begin{equation}
\begin{aligned}
    \mathrm{PE}_{(t_i,2k)} = \sin\!\Bigl(\tfrac{t_i}{10000^{2k/d_{\mathrm{pos}}}}\Bigr), \\
    \mathrm{PE}_{(t_i,2k+1)} = \cos\!\Bigl(\tfrac{t_i}{10000^{2k/d_{\mathrm{pos}}}}\Bigr).
\end{aligned}
\end{equation}
For node pair $(i,j)$, we concatenate both encodings and reduce dimensionality to $d_2 = \lfloor d_t/2\rfloor$ dimensions.

After concatenating the above features, we apply linear transformation, LayerNorm, ReLU, and Dropout for fusion, yielding the final temporal embedding $\mathbf{e}_{\mathrm{time}}^{(ij)} \in \mathbb{R}^{d_t}$.

\subsubsection{Temporal-Aware Triple Attention}
\label{subsec:triple-attention}

To comprehensively model topological structures, temporal dependencies, and global context, this paper computes three types of attention in parallel within each GAT layer:

\begin{itemize}
  \item \textbf{Structural attention} $\alpha_{ij}^{\mathrm{struct}}$: Multi-head dot-product attention based on node features
  \item \textbf{Temporal attention} $\alpha_{ij}^{\mathrm{temp}}$: Incorporating temporal embedding $\mathbf{e}_{\mathrm{time}}^{(ij)}$ into keys/values to capture temporal correlations
  \item \textbf{Global context attention} $\alpha_{ij}^{\mathrm{global}}$: Global normalization of all edge features
\end{itemize}

An adaptive fusion network dynamically computes weights $[w_s,w_t,w_g]$ to obtain the final attention:
\begin{equation}
  \alpha_{ij} = w_s\,\alpha_{ij}^{\mathrm{struct}} + w_t\,\alpha_{ij}^{\mathrm{temp}} + w_g\,\alpha_{ij}^{\mathrm{global}}.
\end{equation}

Finally, multi-head aggregation and message passing are performed to update node representations.

\subsubsection{Cost-Sensitive Loss Optimization}\label{subsec:loss}

In transaction graphs, legitimate transactions constitute approximately 98\% while fraudulent transactions represent only about 2\%. Direct application of standard binary cross-entropy (BCE) biases the model toward predicting the legitimate class. Therefore, this paper employs a weighted BCE loss function:

With positive sample weight $w_{\mathrm{pos}} = N_{\mathrm{neg}}/N_{\mathrm{pos}}$ and negative sample weight $w_{\mathrm{neg}} = 1$, we have:
\begin{equation}
    \mathcal{L}_{\mathrm{WBCE}} = -\bigl[w_{\mathrm{pos}}\,y\,\log(\hat{y}) + w_{\mathrm{neg}}\,(1-y)\,\log(1-\hat{y})\bigr].
\end{equation}

\section{Experiments}

\subsection{Dataset and Experimental Setup}
This study utilizes the transaction subgraph from Elliptic++ \cite{elmougy2023demystifying}, mapping Bitcoin transactions to real entities in licit categories (exchanges, wallet providers, legitimate services) versus illicit categories (scams, malware, terrorist organizations, ransomware). The dataset contains 203,769 nodes and 234,355 edges, with 2\% (4,545) nodes labeled as illicit and 21\% (42,019) as licit, leaving 77\% unlabeled.

Each node contains 183-dimensional features: (1) 94 local features describing transaction properties (input/output quantities, fees); (2) 72 aggregated features from one-hop neighbor statistics (max, min, std, correlations); (3) 17 enhanced features added by Elliptic++. Transactions span 49 discrete time steps with $\approx 2 \text{-week}$ intervals, covering nearly two years.

Considering experimental controllability and computational constraints, we adopt a supervised learning framework using only the 46,564 labeled nodes. These are split 8:1:1 into training, validation, and held-out sets, with the held-out portion reserved but unused in our experiments. Models are trained for 170 epochs using the AdamW optimizer (lr=0.005) with cosine annealing.

\subsection{Baseline Methods}
To comprehensively evaluate the proposed method's performance, we select the following two categories of baseline methods for comparison:

\textbf{Traditional Machine Learning Methods}: (1) Logistic Regression(LR): representative of linear classifiers; (2) Decision Tree(DT): capable of capturing nonlinear decision boundaries; (3) Random Forest(RF): typical representative of ensemble learning methods; (4) XGBoost(XGB): gradient boosting method with excellent performance in numerous competitions. These methods directly use node features for classification without considering graph structure information.

\textbf{GNN Methods}: (1) GCN with BCELoss(GCN): standard graph convolutional network using binary cross-entropy loss; (2) GCN with WeightedBCELoss(GCN-W): GCN with weighted loss function to handle class imbalance. These methods can utilize graph structure information but lack temporal modeling capabilities.

\subsection{Evaluation Metrics}

Considering the specificity of fraud detection tasks and severe dataset imbalance, this paper adopts the following evaluation metrics: AUC (Area Under ROC Curve) as the primary metric, as it is unaffected by class distribution and comprehensively evaluates model performance across different thresholds; Accuracy, Precision, Recall, and F1-Macro as auxiliary metrics, providing multi-dimensional performance assessment.

\subsection{Benchmark Experiments}

Table~\ref{tab6} presents comprehensive performance comparison of various methods on the Elliptic++ dataset. Experimental results demonstrate that the proposed ATGAT method achieves optimal or near-optimal performance across multiple metrics.

\begin{table}[H]
    \centering
    \caption{Performance Comparison on Elliptic++ Dataset}
    \label{tab6}
    \adjustbox{width=\columnwidth}{
    \begin{tabular}{lccccc}
        \toprule
        Model & Accuracy & Precision & Recall & F1-Macro & AUC \\
        \midrule
        LR & 0.9195 & 0.3586 & 0.6556 & 0.7101 & 0.7949 \\
        DT & 0.8872 & 0.2875 & 0.7612 & 0.6774 & 0.8279 \\
        RF & 0.9810 & 0.9708 & 0.6622 & 0.8887 & 0.8305 \\
        XGB & 0.9803 & 0.9358 & 0.6755 & 0.8872 & 0.8364 \\
        GCN & 0.9153 & 0.3421 & 0.6456 & 0.7007 & 0.8144 \\
        GCN-W & 0.8997 & 0.3036 & 0.6887 & 0.6832 & 0.8683 \\
        ATGAT-W & 0.9715 & 0.7924 & 0.6278 & 0.8427 & 0.9130 \\
        \bottomrule
    \end{tabular}
    }
\end{table}

From the experimental results, we observe the following phenomena: (1) Among traditional machine learning methods, XGBoost performs best with an AUC of 0.8364, but still significantly lower than graph neural network methods; (2) Standard GCN achieves an AUC of 0.8144, which improves to 0.8683 with weighted loss, validating the necessity of handling class imbalance; (3) ATGAT achieves an AUC of 0.9130, representing a 9.2\% improvement over the best traditional method XGBoost, 12.1\% over standard GCN, and 5.1\% over weighted GCN.

Notably, while RandomForest excels in Precision (0.9708) and F1-Macro (0.8887), its AUC is only 0.8305, indicating good performance at fixed thresholds but inferior comprehensive performance across different thresholds compared to ATGAT. This phenomenon further validates the rationality of AUC as the primary evaluation metric.

\subsection{Ablation Study}

To comprehensively understand each model component's contribution, we designed systematic ablation experiments. By progressively removing or replacing key modules, we quantitatively assess each component's importance. All ablation experiments were repeated 10 times under identical settings with different random seeds to ensure result stability.

The ablation study includes five model variants: (1) Baseline GAT(B-GAT): standard graph attention network only, without temporal information or enhancement modules; (2) Structural GAT(S-GAT): retaining only structural attention mechanism; (3) Temporal GAT(T-GAT): retaining only temporal attention mechanism; (4) ATGAT with BCELoss(ATGAT): complete triple attention network; (5) ATGAT with WeightedBCELoss(ATGAT-W): ATGAT with added class weighting strategy.

\begin{table}[htbp]
    \centering
    \caption{Ablation Study Results (Mean ± Std Dev)}
    \label{tab:ablation}
    \adjustbox{width=\columnwidth}{
    \begin{tabular}{lccccc}
        \toprule
        Model Variant & Accuracy & Precision & Recall & F1-Macro & AUC \\
        \midrule
        B-GAT & 0.9110±0.0313 & 0.2984±0.1177 & 0.5844±0.2100 & 0.6712±0.0722 & 0.8295±0.0260 \\
        S-GAT & 0.9714±0.0015 & \textbf{0.7982±0.0268} & 0.6166±0.0044 & 0.8403±0.0063 & 0.8993±0.0046 \\
        T-GAT & 0.9713±0.0011 & 0.7948±0.0208 & 0.6189±0.0076 & 0.8403±0.0048 & 0.8989±0.0078 \\
        ATGAT & 0.9709±0.0011 & 0.7806±0.0201 & \textbf{0.6298±0.0078} & 0.8409±0.0049 & 0.9111±0.0043 \\
        ATGAT-W & \textbf{0.9715±0.0012} & 0.7924±0.0235 & 0.6278±0.0087 & \textbf{0.8427±0.0046} & \textbf{0.9130±0.0055} \\
        \bottomrule
    \end{tabular}
    }
\end{table}

Table~\ref{tab:ablation} reveals key findings: Baseline GAT performs poorly (AUC=0.8295±0.0260) with high variance, demonstrating standard GAT's inadequacy for imbalanced tasks. Single attention mechanisms (Structural/Temporal GAT) improve AUC to ~0.899, validating their effectiveness. Complete ATGAT achieves AUC=0.9111±0.0043 with the highest Recall (0.6298±0.0078), confirming triple attention synergy. ATGAT-W further improves to AUC=0.9130±0.0055 through class weighting, validating the cost-sensitive strategy.

\subsection{Effect Size Analysis}

To comprehensively evaluate the practical significance of model improvements, we calculated Cohen's d effect sizes to quantify performance differences between model variants.

\begin{table}[htbp]
    \centering
    \caption{Cohen's d Effect Sizes for AUC Differences}
    \label{tab:significance}
    \adjustbox{width=\columnwidth}{
    \begin{tabular}{lccccc}
        \toprule
        & B-GAT & S-GAT & T-GAT & ATGAT & ATGAT-W \\
        \midrule
        B-GAT & 0.000 & -3.734 & -3.616 & -4.378 & -4.442 \\
        S-GAT & 3.734 & 0.000 & 0.048 & -2.653 & -2.704 \\
        T-GAT & 3.616 & -0.048 & 0.000 & -1.929 & -2.081 \\
        ATGAT & 4.378 & 2.653 & 1.929 & 0.000 & -0.381 \\
        ATGAT-W & 4.442 & 2.704 & 2.081 & 0.381 & 0.000 \\
        \bottomrule
    \end{tabular}
    }
\end{table}

\noindent\textit{According to Cohen's criteria, $|d| < 0.2$ indicates negligible effect; 0.2-0.5 for small; 0.5-0.8 for medium; $> 0.8$ for large.}

\vspace{1pt}

To quantify the practical significance of the model improvement, Cohen's d effect sizes were calculated for the key comparisons.The results on AUC were 4.442 for Baseline GAT vs. ATGAT-Weighted, 2.653 for Structural GAT vs. ATGAT, and 0.381.These results suggest that the improvements from Baseline GAT to the full model, and from single to triple attention mechanisms have of extreme practical significance, while the category-weighted strategy, although it brings relatively small enhancements, still has practical applications.

\section{Conclusion}

This paper presents an Augmented Temporal-aware Graph Attention Network (ATGAT) for cryptocurrency transaction fraud detection. The proposed method addresses the limitations of existing approaches through three key modules: First, the designed advanced temporal embedding module effectively captures dynamic temporal patterns in transactions by fusing multi-scale time difference features with periodic position encoding; then, the triple attention mechanism learns node representations from structural, temporal, and global dimensions, significantly enhancing model expressiveness; finally, the weighted loss strategy systematically addresses the extreme class imbalance problem in fraud detection.

Experimental evaluation on the Elliptic++ cryptocurrency dataset fully validates the effectiveness of the proposed method. ATGAT achieves an AUC of 0.9130, representing a 9.2\% improvement over the best traditional method XGBoost and a 12.1\% improvement over standard graph neural network GCN. Systematic ablation experiments demonstrate that temporal embedding, triple attention mechanism, and weighting strategy all contribute significantly to performance improvement. Notably, Cohen's d effect size analysis reveals that model improvements possess extremely large statistical significance and practical application value.

Despite the achievements of this research, several limitations and areas remain for future investigation. First, constrained by computational resources, this study employs empirically-based fixed hyperparameter settings without a systematic hyperparameter search, using a training-validation two-stage evaluation that may have further optimization potential. Second, the dataset used focuses on the Bitcoin transaction domain with limited data diversity and breadth, potentially affecting model generalization to other financial transaction scenarios. Additionally, while ATGAT excels in AUC performance, it does not achieve state-of-the-art results on some threshold-based metrics, which could be optimized through threshold adjustment to fully realize the model's potential. Through these efforts, we will further advance financial fraud detection technology and provide more effective technical support for financial security and risk control.

\bibliographystyle{IEEEtran}
\bibliography{IEEEabrv,cited}
\end{document}